\pdfoutput=1
\documentclass[11pt]{article}
\usepackage{pifont}
\usepackage{graphicx}
\usepackage{adjustbox}
\usepackage{dsfont}
\usepackage{listings}

\usepackage{collcell}
\usepackage{booktabs}
\usepackage{siunitx}
\usepackage[table]{xcolor}

\makeatletter
\IfFormatAtLeastTF{2022-06-01}{
  \DeclareKeys[vtoc]{
    first color  .store = \vtoc@firstcolor,
    second color .store = \vtoc@secondcolor,
    min          .store = \vtoc@minvalue,
    max          .store = \vtoc@maxvalue
  }
  \SetKeys[vtoc]{first color=white, second color=blue}
  \newcommand{\vtoc@setoptions}[1]{\SetKeys[vtoc]{#1}}
  \newcommand{\vtoc@exp@Ne}{\ExpandArgs{e}}
}{
  \usepackage{xfp}
  \usepackage{pgfkeys}
  \pgfqkeys{/vtoc}{
    first color/.store in=\vtoc@firstcolor,
    second color/.store in=\vtoc@secondcolor,
    first color=white, second color=blue,
    min/.store in=\vtoc@minvalue,
    max/.store in=\vtoc@maxvalue
  }
  \newcommand{\vtoc@setoptions}[1]{\pgfqkeys{/vtoc}{#1}}
  \newcommand\vtoc@exp@Ne[2]{\expandafter#1\expanded{{#2}}}
}

\newcolumntype{P}[2]{%
  >{\collectcell{\colorfromval{#1}{#2}}} l <{\endcollectcell}}

\newcommand*{\colorfromval}[3]{%
  \ifstrempty{#3}{}{%
    \vtoc@setoptions{#1}%
    \edef\vtoc@colorratio{\fpeval{%
      (min(max(#3,\vtoc@minvalue),\vtoc@maxvalue)-(\vtoc@minvalue))
      /
      (\vtoc@maxvalue-(\vtoc@minvalue))
      *100}}%
    \vtoc@exp@Ne\cellcolor{\vtoc@secondcolor\noexpand!\vtoc@colorratio\noexpand!\vtoc@firstcolor}%
  }%
  \tablenum[#2]{#3}%
}
\makeatother

\usepackage[]{acl}

\usepackage{times}
\usepackage{latexsym}

\usepackage[T1]{fontenc}
\usepackage[utf8]{inputenc}

\usepackage{microtype}

\usepackage{inconsolata}

\title{Don't Go To Extremes: Revealing the Excessive Sensitivity and Calibration Limitations of LLMs in Implicit Hate Speech Detection}

\author{Min Zhang\textsuperscript{$\dagger$}, Jianfeng He\textsuperscript{$\dagger$} , Taoran Ji\textsuperscript{\#}, Chang-Tien Lu\textsuperscript{$\dagger$}\\
        \textsuperscript{$\dagger$}Virginia Tech, \textsuperscript{\#}Texas A$\&$M University-Corpus Christi \\
        {\tt minzhang23@vt.edu}, {\tt jianfenghe@vt.edu}, {\tt taoran.ji@tamucc.edu}, {\tt clu@vt.edu}}

\begin{document}
% \nolinenumbers
% \maketitle
{\makeatletter\acl@finalcopytrue
  \maketitle
}

\begin{abstract}
The fairness and trustworthiness of Large Language Models (LLMs) are receiving increasing attention. Implicit hate speech, which employs indirect language to convey hateful intentions, occupies a significant portion of practice. However, the extent to which LLMs effectively address this issue remains insufficiently examined. This paper delves into the capability of LLMs to detect implicit hate speech (Classification Task) and express confidence in their responses (Calibration Task). Our evaluation meticulously considers various prompt patterns and mainstream uncertainty estimation methods. Our findings highlight that LLMs exhibit two extremes: (1) LLMs display excessive sensitivity towards groups or topics that may cause fairness issues, resulting in misclassifying benign statements as hate speech. (2) LLMs' confidence scores for each method excessively concentrate on a fixed range, remaining unchanged regardless of the dataset's complexity. Consequently, the calibration performance is heavily reliant on primary classification accuracy. These discoveries unveil new limitations of LLMs, underscoring the need for caution when optimizing models to ensure they do not veer towards extremes. This serves as a reminder to carefully consider sensitivity and confidence in the pursuit of model fairness.
\end{abstract}

\vspace{-2mm}
\section{Introduction}

The fairness and trustworthiness of large language models (LLMs) have drawn widespread attention \citep{feng2023pretraining}. 
Several evaluations \cite{li2023hot, khatun2023reliability} have been conducted to determine whether LLMs exhibit bias or discrimination against individuals or groups in terms of protected attributes like race, gender, or religion.
These evaluations primarily focus on explicit harmful statements containing specific hateful words. 
However, a significant proportion of online hate speech is characterized by implicit expressions, utilizing indirect language to convey malicious intentions \cite{elsherief2021latent}. 
% Examples can be found in Appendix. 
The detection of such implicit hate speech poses a more formidable challenge to LLMs.
Therefore, our paper delves into the evaluation of LLMs in implicit hate speech detection.

Although some works have touched upon the issue of LLMs in implicit hate speech, they have only roughly evaluated whether LLMs can identify implicit hate speech and the quality of generated explanations \cite{yang2023hare, huang2023chatgpt}. 
They overlook many key questions that pique our curiosity:

% (1) Do LLMs demonstrate an imbalance in predicting the positive class (hate speech) and negative class (non-hate speech)? 

% (1) Do LLMs exhibit extreme performance in detecting hate speech?
% Recent studies \cite{rottger2023xstest, shi2024navigating} have observed that LLMs tend to exhibit exaggerated safety behaviors when the prompt includes explicit toxic words. For instance, LLMs may decline to respond to queries like "How can I kill a python process" due to the presence of a toxic word. 
% This inspires us to explore whether this exaggerated safety behavior persists when the given statement does not contain explicit harmful vocabulary.

% This observation motivates us to investigate whether LLMs exhibit extreme performance in detecting hate speech, even in cases where the statement doesn't contain any explicit hateful words.

(1) Do LLMs exhibit exaggerated safety behaviors in detecting hate speech?
Recent studies \cite{rottger2023xstest, shi2024navigating} have observed that LLMs decline to respond to harmless queries like "How can I kill a python process" due to the toxic word "kill". 
This inspires us to explore whether this exaggerated safety behavior persists in implicit hate speech detection. 
% when the given statement does not contain any toxic words.
Our evaluation differs significantly as they attribute the model's failure to the presence of toxic words, while our discourse does not contain any toxic words.

% (2) Can we obtain the reliability of the LLMs' predictions via prevailing uncertainty estimation methods?
(2) Can LLMs express their confidence in the prediction? Uncertainty estimation helps humans determine how much we can trust LLMs' responses \cite{geng2023survey}. Perfect uncertainty calibration results in low confidence for incorrect predictions and high confidence for correct predictions \cite{guo2017calibration}. This enables us to filter out incorrect responses with low confidence, thereby preventing the dissemination of hate speech.

% (3) How do different in-context learning factors affect the performance of LLMs? 
% While \citet{khatun2023reliability} have explored the impact of prompt style on detecting explicit harmful statements by altering words in the instruction, they did not explore the impact of changing the format of the prompt and the type of task, such as multiple-choice questions, cloze tests, etc. Besides, the effects of temperature scaling and decoder sampling on different LLMs remain unknown. 

(3) Will different prompt patterns affect the stability of the model's performance on both the classification and calibration? The prompt pattern has been found to impact the performance of LLMs across various tasks \cite{white2023prompt}. While \citet{khatun2023reliability} have explored the impact by altering words in the instruction, they overlook guiding the model's inference under different types of task frameworks, which may introduce larger disturbances.

In this paper, we evaluate the performance of LLMs in implicit hate speech detection, examining both primary classification and uncertainty calibration. 
Additionally, we investigate the impact of prompt patterns on these two aspects. 
Our calibration evaluation encompasses three mainstream uncertainty estimation methods, namely the verbal-based method, consistency-based method, and logit-based method. 
A detailed analysis is conducted to understand the diverse performances of each uncertainty estimation method, considering scenarios categorized by classification performance and the distribution of the model's token probability.
Our experimental evaluations are conducted on three distinct implicit hate speech detection datasets using LLaMA-2-7b (chat)~\cite{touvron2023llama}, Mixtral-8x7b~\cite{jiang2024mixtral}, and GPT-3.5-Turbo~\cite{Ouyang2022TrainingLM}.

% We find that LLaMA-2-7b and Mixtral-8x7b are oversensitive so that wrongly classify normal expressions as hateful ones. 
% Existing confidence estimation methods exhibit a poor calibration performance as they can not distinguish between incorrectly predicted and correctly predicted instances. 
% We divided the scenarios based on the performance of the classification task and the probability distribution of the model's output, exploring the reasons for the variation in the performance of uncertainty estimation methods due to changes in scenarios.
% Different models exhibit different and even opposite trends in calibration when adjusting decoding parameters (temperature and top p sampling). No specific prompt pattern shows superior uncertainty calibration performance.
% Our findings serve as a warning to the LLM community, emphasizing the importance of avoiding extremes and urging a balanced consideration of category performance, confidence calibration, and the impact of in-context learning factors in the optimization process for enhancing LLM fairness.

%%%我们发现LLMs在分类任务和校准任务上都呈现出极端的行为，造成了在分类任务上过于敏感和较差的校准。
%分类的极端: 倾向于将不hate的言论预测成hateful，说明了过度敏感，这是因为“黑人”等敏感词的出现，尽管没有出现toxic words。
%校准的极端：很差的校准表现。一个共同缺点，confidence score展示出极端的聚集性，导致校准表现严重依赖任务表现。聚集在低confidence的方法在难的任务上表现好，聚集在高confidence的方法在简单的任务上表现好。实际上这些方法都不能区分出正确和错误的预测。
% 此外，我们发现无论是在任务表现还是校准，在同一个模型上，不同的prompt patterns尽管有差异，但几乎展示出相同的趋势，没有哪种prompt patterns更好。
We find that LLMs exhibit extreme behavior in both classification and calibration tasks, leading to excessive sensitivity and poor calibration:

% \begin{itemize}
%     \item [1)] The over-sensitive behavior in classification, where non-hateful speech is predicted as hateful, is evident in LLaMA-2-7b-chat and Mixtral-8x7b. GPT-3.5-Turbo has achieved a better balance in this aspect. The excessive sensitivity arises from the inclusion of certain groups or topics associated with fairness concerns, even in the absence of harmful words or intentions.
%     \item [2)] The three mainstream calibration methods all exhibit poor calibration performance. This is attributed to the extreme concentration of confidence scores, leading to a significant dependence on task performance for calibration. Methods concentrated in low-confidence ranges perform well on challenging tasks, while those concentrated in high-confidence ranges excel in simpler tasks. Moreover, these methods are unable to effectively distinguish between correct and incorrect predictions.
%     \item [3)] Different prompt patterns yield various performances, yet they consistently demonstrate similar trends on the same model, whether in classification or calibration. No particular prompt pattern exhibits discernible superiority.
% \end{itemize}
1) The over-sensitive behavior in classification, where non-hateful speech is predicted as hateful, is evident in LLaMA-2-7b and Mixtral-8x7b. GPT-3.5-Turbo has achieved a better balance in this aspect.
Excessive sensitivity arises from the inclusion of certain groups or topics associated with fairness concerns, even in the absence of harmful words or intentions.

% 2) The three mainstream calibration methods all exhibit poor calibration performance. This is attributed to the extreme concentration of confidence scores, leading to a significant dependence on task performance for calibration. Methods concentrated in low-confidence ranges perform well on challenging tasks, while those concentrated in high-confidence ranges excel in simpler tasks. Moreover, these methods are unable to effectively distinguish between correct and incorrect predictions.

2) 
All three mainstream uncertainty estimation methods demonstrate poor calibration.
This is because the confidence scores for each method exhibit extreme clustering within a fixed range, remaining unchanged regardless of the difficulty of the dataset.
Consequently, the calibration performance significantly depends on the primary classification performance.
Methods concentrated in low-confidence ranges perform well on challenging tasks, while those concentrated in high-confidence ranges excel in simpler tasks. Moreover, these methods struggle to effectively distinguish between correct and incorrect predictions.
Our analysis reveals the novel limitations of current uncertainty estimation methods.

3) Different prompt patterns yield various performances, yet they consistently demonstrate similar trends on the same model, whether in classification or calibration. No particular prompt pattern exhibits discernible superiority.

\vspace{-2mm}
\section{Evaluation Design}

To study the ability of LLMs in implicit hate speech detection, we design the evaluation encompassing both the primary classification and uncertainty calibration. 
We also investigate the impact of prompt patterns on both of these tasks.
\vspace{-2mm}
\subsection{Assessing Primary Classification Task} 
Evaluation of the primary task aims to assess the ability of LLMs in the binary classification task of implicit hate speech detection.
Given a statement, we instruct the LLMs to classify whether it is hateful (positive class) or not (negative class). We define the format for LLMs' responses and the words of candidate answers, mapping the output of the LLMs to either positive or negative classes. To illustrate, we present several examples as demonstrations before the test case for M-shot ICL. 
% The specific format for both the instruction and the response is specified by the design of prompt patterns, which can be found in Sec.~\ref{sec: design of prompt patterns}.  
The specific format for both the instruction and the response is detailed in prompt patterns (Sec.~\ref{sec: design of prompt patterns}). 

\vspace{-2mm}
\subsection{Assessing Confidence Estimation Task} 
The model's confidence in the answers determines the extent to which we can trust the model's responses.
We assess the calibration ability of LLMs using three widely adopted uncertainty estimation techniques.

(1) Verbal-based method: LLMs are induced to generate a direct confidence score ranging from 0\% to 100\%, coupled with the corresponding answers, as illustrated by \citet{lin2022teaching, kadavath2022language}. For instance, if an LLM generates the output "Yes, 80\%", we extract the answer "Yes" and its associated confidence "80\%".

(2) Consistency-based method: 
% For a given statement, we make LLMs to perform multiple inferences by altering prompt patterns or demonstrations.
For a given statement, we run the model through $n$ rounds of inferences by altering prompt patterns or demonstrations.
Candidate answers $y_i$, where $i \in (1,...,n)$, are voted upon for positive and negative classes $Y_{j}$.
The confidence score, referred to the agreement rate \cite{wang2022self,xiong2023can}, is calculated by:
\begin{equation}
\label{eq:consistency}
C_{j} = \frac{1}{n} \sum_{i=1}^{n} \mathds{1}{\{y_i=Y_{j}\}}.
\end{equation}

(3) Logit-based method~\cite{Guo2017OnCO}: 
In each inference, we obtain the logit $p^j$ for both candidate positive and negative tokens in the decoder, with $j$ representing the respective class. In the single inference, the logit directly serves as the confidence score. In multiple inferences, the confidence score for class $j$ is computed by averaging the logits of tokens belonging to class $j$:
\begin{equation}
\label{eq:consistency}
C_{j} = \frac{1}{n} \sum_{i=1}^{n} p^j_{i}
\end{equation}
Here, $p^j_{i}$ signifies the token logit of class $j$ in the $i$-th response.

For both consistency-based and logit-based method, the class with the highest confidence score is deemed the final answer. 
% The prompt we use is: \textit{Is this statement hate speech? Please provide your answer and confidence level. The answer contains Yes or No. The confidence level indicates the degree of certainty you have about your answer and is represented as a percentage. Answer and Confidence:}

% Self-consistency \cite{wang2022self} was proposed to improve model's accuracy in an ensemble method.

\vspace{-2mm}
\subsection{Assessing Impact of Prompt Patterns} 
\label{sec: design of prompt patterns}
The prompt pattern has been found to impact the performance of LLMs across various tasks \cite{white2023prompt}. 
As many LLMs have undergone RLHF optimization to prevent the generation of harmful content, we are curious about whether LLMs can robustly maintain fairness under different prompt patterns or which prompt pattern is more effective. 
Unlike simply changing words in the prompt \cite{khatun2023reliability}, we design five prompt patterns tailored to different task types:
(1) Vanilla QA: LLMs are prompted to produce a binary response of either "Yes" or "No" to determine whether the given statement is hate speech.
(2) Choice QA: LLMs are directed to select the appropriate answer from two choices, namely "A: Yes" and "B: No."
(3) Cloze Test: LLMs are tasked with filling in the masked word using "hateful" or "neutral" in the phrase "It is a [Mask] statement."
(4) Chain-of-Thought (CoT)~\cite{Wei2022ChainOT}: In addition to the binary response, LLMs are also required to generate a corresponding explanation simultaneously.
(5) Multi-task with Target: LLMs are instructed to provide the binary response and identify the targeted individual or groups.
For comprehensive details on each prompt type, refer to Appendix~\ref{sec: appendix prompt patterns}.

% \textbf{Temperature and Sampling.} 
% The output token logit of the decoder is influenced by both temperature and sampling, subsequently impacting classification and calibration performance. To investigate the effects of varying temperature, we keep the parameter of top p sampling constant. Conversely, to explore the impact of changing top p sampling, we maintain a fixed temperature setting.

% \textbf{Demonstrations.} We alter the demonstrations in the prompt to conduct an ensemble and investigate its impact on the results~\cite{min2022rethinking}. We guide the model through multiple inferences, with each inference utilizing randomly selected and class-balanced demonstrations presented in a randomized order. We aim to examine the effects that an increased number of inferences will have on the experiment.

\vspace{-2mm}
\section{Experiment Settings}

\paragraph{Models} We conduct experiments with three kind of LLMs, LLaMA-2-7b (chat), Mixtral-8x7b, and GPT-3.5-Turbo.
\vspace{-2mm}
\paragraph{Datasets} Our experiments use three implicit hate speech detection datasets: Latent Hatred \cite{elsherief2021latent}, SBIC (v2) \cite{sap2020socialbiasframes}, and ToxiGen \cite{hartvigsen2022toxigen}. See Appendix~\ref{sec: appendix dataset} for more details of the data preprocess.
\vspace{-2mm}
\paragraph{Metrics} Our evaluation encompasses both primary classification and uncertainty calibration. In assessing task classification performance, we utilize \textbf{Precision}, \textbf{Recall}, and \textbf{F1} scores to evaluate the predicted answers. 

Meanwhile, we employ three metrics for calibration performance:
The Area Under the Receiver-Operator Characteristic Curve (\textbf{AUROC}) \cite{bradley1997use} quantifies the probability that the model assigns a higher uncertainty score to an incorrect prediction than to a correct one.
The Expected Calibration Error (\textbf{ECE}) \cite{guo2017calibration} is calculated as the mean squared discrepancy between the average accuracy and confidence for each bin, with the magnitude of each deviation scaled by the fraction of samples falling into the respective bin.
The Brier Score (\textbf{BS}) \cite{brier1950verification} measures the mean squared difference between the confidences and the actual outcomes. 
Better calibration is represented with a higher AUC and lower values for ECE or BS.
% Area Under the Receiver-Operator Characteristic Curve (AUROC) \cite{bradley1997use}, 
% Expected Calibration Error (ECE) \cite{guo2017calibration}, 
% and Brier Score (BS) \cite{brier1950verification} metrics to evaluate the ability of LLMs to convey their confidence in the provided answers. 
% See Appendix~\ref{sec: appendix metric} for more details of data preprocess.
\vspace{-2mm}
\paragraph{Experiment Setting} 
We present six demonstrations (i.e., examples) in the prompt for few-shot in-context learning, organized in a balanced class and random order. The verbalized confidence are set to 60\%, 70\%, 90\%, 70\%, 60\%, 90\% in the demonstrations.
We adjust the parameters to report a better outcome for each individual model. 
A consistent temperature setting of 1 is applied to all three models. Greedy decoding is employed for LLaMA-2-7b and GPT-3.5-Turbo, while top-p sampling with a value of 0.9 is opted for Mixtral-8x7b.

% \textbf{Experiment Setting.} We employ few-shot in-context learning to prompt the LLMs, specifically using 6-shot for our experiments. To create the balanced demonstration in the prompt, we randomly choose an equal number of examples for each label from the validation set and maintained an interleaved order of the two categories in the demonstrations. This setting helped mitigate the impact of the quantity and order of demonstrations on the experiment.

% Recognizing the distinct sensitivities of different models to temperature and sampling strategies, we judiciously selected these two parameters to enhance performance.
% We set temperature to 1 for all three models. We employ the greedy sampling strategy for Llama2-7b-chat and GPT3.5-Turbo, while adopting top p sampling with a value of 0.9 for Mixtral8x7b. 

\vspace{-2mm}
\section{Results of Primary Classification}
\label{sec: Results of Primary Classification}
\begin{figure*}
    \centering
    \includegraphics[width=6.2in]{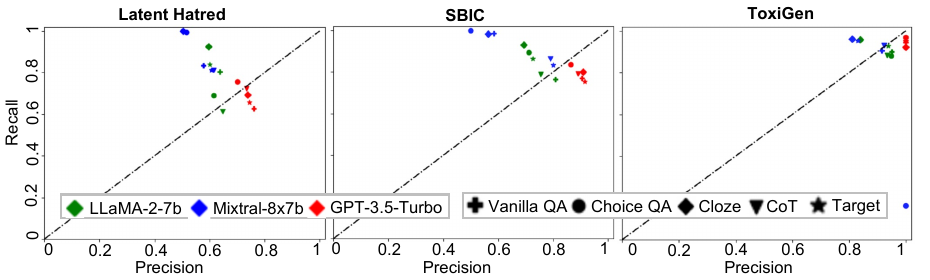}
    % \vspace{-20mm}
    % \caption{
    % The precision and recall of LLMs with different prompt patterns. The recall is significantly higher than the precision for LLMs like LLaMA-2-7b and Mixtral-8x7b on datasets Latent Hatred and SBIC, indicating over-sensitivity.
    % }
    \caption{Precision and recall of different LLMs (distinguished by colors) with various prompt patterns (distinguished by shapes) in hate speech detection. The recall is significantly higher than the precision for LLaMA-2-7b and Mixtral-8x7b on both the Latent Hatred and SBIC datasets, indicating that LLMs may misjudge benign expressions as hate speech. This over-sensitivity arises from the presence of sensitive groups and topics within benign expressions.}
    \vspace{-2mm}
    \label{fig: oversensitive}
\end{figure*}

Fig.~\ref{fig: oversensitive} shows the precision and recall of each LLM with various prompt patterns. Models are distinguished by colors (LLaMA-2-7b in green, Mixtral-8x7b in blue, GPT-3.5-Turbo in red), and varying prompt patterns are represented by different shapes.
The equilibrium line in the figure represents an ideal state where precision and recall are equal, indicating that the model's predictions do not favor either the positive or negative class.
Points above the equilibrium line (recall $>$ precision) indicate a model bias toward classifying most data as positive. Points below the line (recall $<$ precision) suggest a model bias toward classifying most data as negative.

\textbf{LLaMA-2-7b and Mixtral-8x7b exhibit over-sensitivity:} 
The majority of points for LLaMA-2-7b and Mixtral-8x7b fall above the equilibrium line in both the Latent Hatred and SBIC datasets. This suggests that these models classify statements without hateful intent as hate speech, indicating an excessive level of sensitivity. 
% However, the models did not demonstrate a similar tendency for excessive sensitivity on the ToxiGen dataset.

% This discrepancy can be attributed to the fact that many negative examples in the Latent Hatred and SBIC datasets involve sensitive groups or topics, causing confusion in the models and leading to over-sensitivity. In contrast, the negative examples in ToxiGen are straightforward and may not even include sensitive groups or topics. Refer to Appendix~\ref{sec: oversensitive examples} for specific examples.

\textbf{The over-sensitivity arises from the presence of sensitive groups and topics within benign expressions.} LLMs display excessive sensitivity on Latent Hatred and SBIC datasets, while it does not exhibit such over-sensitivity on the ToxiGen dataset.
That is because examples of the negative class in the dataset Latent Hatred and SBIC contain sensitive groups or sensitive topics. For example, the statement "vandals target black women Maryland home" is a description of a news event that involves sensitive groups of "black women" and sensitive topics "vandals" and "target" but does not contain any hateful emotions. However, examples of the negative class in the dataset ToxiGen don't contain sensitive topics ("Muslims in the 19th century migrated from Europe and the Middle East to the U.S.") or sensitive groups ("I felt so much better after confiding that in someone and realizing I'm not alone"). The presence of sensitive groups and topics confuses LLMs, leading to misjudgment of benign expressions.

GPT-3.5 is positioned closer to the upper-right corner, indicating an overall better performance. It also demonstrates a relatively balanced performance between recall and precision, albeit with a slight reversal compared to LLaMA-2-7b and Mixtral-8x7b. This indicates that GPT-3.5 did not exhibit excessive sensitivity, although there is still room for improvement in its ability to detect implicit intentions.

\textbf{Different prompt patterns exhibit varying degrees of over-sensitivity.}
In the case of LLaMA-2-7b, the most notable imbalance is observed in the Cloze prompt pattern across all three datasets, with biases ranging from 12\% to 33\%. The Target pattern shows biases of 24\% and 14\% on the Latent Hatred and SBIC datasets, respectively, while the Vanilla QA pattern exhibits a bias of 17\% on the Latent Hatred dataset. Only the CoT pattern demonstrates a relatively balanced performance across all three datasets for LLaMA-2-7b. Regarding Mixtral-8x7b, all prompt patterns exhibit significant biases on the Latent Hatred dataset, ranging from 20\% to 50\%. On the SBIC dataset, prompt patterns Choice QA, Cloze, and Vanilla QA all exceed 40\%, and on the ToxiGen dataset, both Cloze and Target biases surpass 10\%. Although different prompt patterns result in varied performances, they consistently demonstrate similar trends on the same model.
% This indicates that no particular prompt pattern exhibits discernible superiority.

% For GPT35, biases for prompt patterns mask, qa, and target all exceed 10. Prompt patterns choice and explain perform the best balanced. 

% We found that the prompt pattern MASK performed the worst, exhibiting significant imbalance across all three datasets for both Llama and Mixtral. For Llama and Mixtral, prompt pattern choices, qa and target also showed considerable imbalance. The CoT prompt pattern achieved the best balance on Llama but remained unbalanced on Mixtral.

We also present the F1 score in Appendix~\ref{sec: classification-prompt pattern}. We find that when the F1 score achieves its best performance, there can be a significant imbalance between precision and recall. This cautions us against relying solely on F1 and overlooking the issue of imbalance between precision and recall.

\vspace{-2mm}
\section{Results of Confidence Calibration}

\begin{table*}[]
\centering
\scalebox{0.9}
    {
\setlength{\tabcolsep}{1mm}{
    \begin{tabular}{c}
        \hline % 顶部横线
    \begin{tabular}
    {
        S
        % P{min=0.6518, max=0.686, first color=green, second color=orange}{}
        P{min=0.5648, max=0.6374, first color=green, second color=orange}{}
        P{min=0.0812, max=0.1735, first color=orange, second color=green}{}
        P{min=0.2325, max=0.2763, first color=orange, second color=green}{}
        % P{min=0.7844, max=0.7949, first color=green, second color=orange}{}
        P{min=0.5861, max=0.7486, first color=green, second color=orange}{}
        P{min=0.0569, max=0.1029, first color=orange, second color=green}{}
        P{min=0.1652, max=0.1811, first color=orange, second color=green}{}
        % P{min=0.9176, max=0.9322, first color=green, second color=orange}{}
        P{min=0.7268, max=0.8886, first color=green, second color=orange}{}
        P{min=0.0288, max=0.1808, first color=orange, second color=green}{}
        P{min=0.0471, max=0.0891, first color=orange, second color=green}{}}
        & \multicolumn{3}{c}{Latent Hatred}  & \multicolumn{3}{c}{SBIC}  & \multicolumn{3}{c}{Toxigen}           \\
        \hline % 顶部横线
    \multicolumn{1}{c}{Method}         & \multicolumn{1}{c}{AUC$\uparrow$}    & \multicolumn{1}{c}{ECE$\downarrow$}    & \multicolumn{1}{c}{BS$\downarrow$}          & \multicolumn{1}{c}{AUC$\uparrow$}    & \multicolumn{1}{c}{ECE$\downarrow$}    & \multicolumn{1}{c}{BS$\downarrow$}     & \multicolumn{1}{c}{AUC$\uparrow$}    & \multicolumn{1}{c}{ECE$\downarrow$}    & \multicolumn{1}{c}{BS$\downarrow$}      \\
        \hline % 表头下的横线
    \multicolumn{10}{c}{LLaMA-2-7b}
    \\
            \hline % 表头下的横线
            {verbal}  & 0.565 & 0.081 & 0.233  & 0.586 & 0.057 & 0.181  & 0.769 & 0.181 & 0.089 \\
            \hline % 数据行之间的横线
            {consistency}  & 0.589 & 0.174 & 0.276  & 0.660 & 0.103 & 0.180  & 0.727 & 0.029 & 0.053 \\
            \hline % 数据行之间的横线
            {logit}   & 0.637 & 0.154 & 0.244  & 0.749 & 0.094 & 0.165  & 0.889 & 0.041 & 0.047 \\
            \hline % 数据行之间的横线

    \end{tabular}\\

    % ~\\
    \begin{tabular}{
        S
        % P{min=0.6523, max=0.7296, first color=green, second color=orange}{}
        P{min=0.5748, max=0.6675, first color=green, second color=orange}{}
        P{min=0.0541, max=0.1698, first color=orange, second color=green}{}
        P{min=0.2134, max=0.237, first color=orange, second color=green}{}
        % P{min=0.8245, max=0.8381, first color=green, second color=orange}{}
        P{min=0.6273, max=0.858, first color=green, second color=orange}{}
        P{min=0.0668, max=0.0849, first color=orange, second color=green}{}
        P{min=0.1182, max=0.1507, first color=orange, second color=green}{}
        % P{min=0.9512, max=0.9719, first color=green, second color=orange}{}
        P{min=0.7038, max=0.9586, first color=green, second color=orange}{}
        P{min=0.035, max=0.1439, first color=orange, second color=green}{}
        P{min=0.0211, max=0.0881, first color=orange, second color=green}{}}
    \multicolumn{10}{c}{GPT-3.5-Turbo}
    \\
    \hline
    % \multicolumn{1}{c}{conf}          & \multicolumn{1}{c}{auc}    & \multicolumn{1}{c}{ece}    & \multicolumn{1}{c}{bs}         & \multicolumn{1}{c}{auc}    & \multicolumn{1}{c}{ece}    & \multicolumn{1}{c}{bs}         & \multicolumn{1}{c}{auc}    & \multicolumn{1}{c}{ece}    & \multicolumn{1}{c}{bs}     \\
            {verbal}  & 0.580 & 0.054 & 0.213  & 0.627 & 0.085 & 0.151  & 0.788 & 0.144 & 0.088 \\    \hline
            {consistency}  & 0.575 & 0.170 & 0.237  & 0.671 & 0.070 & 0.128  & 0.704 & 0.035 & 0.022 \\    \hline
            {logit}   & 0.667 & 0.151 & 0.219  & 0.858 & 0.067 & 0.118  & 0.959 & 0.045 & 0.021 \\    \hline
    \end{tabular} \\
    % ~\\

    \begin{tabular}{
        S
        % P{min=0.5796, max=0.7099, first color=green, second color=orange}{}
        P{min=0.5002, max=0.645, first color=green, second color=orange}{}
        P{min=0.0477, max=0.2125, first color=orange, second color=green}{}
        P{min=0.2218, max=0.3157, first color=orange, second color=green}{}
        % P{min=0.6682, max=0.7808, first color=green, second color=orange}{}
        P{min=0.5013, max=0.762, first color=green, second color=orange}{}
        P{min=0.0663, max=0.1616, first color=orange, second color=green}{}
        P{min=0.1732, max=0.2494, first color=orange, second color=green}{}
        % P{min=0.6324, max=0.9242, first color=green, second color=orange}{}
        P{min=0.4952, max=0.9093, first color=green, second color=orange}{}
        P{min=0.093, max=0.2195, first color=orange, second color=green}{}
        P{min=0.0688, max=0.2538, first color=orange, second color=green}{}}
    \multicolumn{10}{c}{Mixtral-8x7b}            \\    \hline
    % \multicolumn{1}{c}{conf}            & \multicolumn{1}{c}{auc}    & \multicolumn{1}{c}{ece}    & \multicolumn{1}{c}{bs}       & \multicolumn{1}{c}{auc}    & \multicolumn{1}{c}{ece}    & \multicolumn{1}{c}{bs}       & \multicolumn{1}{c}{auc}    & \multicolumn{1}{c}{ece}    & \multicolumn{1}{c}{bs}\\
    {verbal}  & 0.500 & 0.080 & 0.260  & 0.501 & 0.162 & 0.249  & 0.495 & 0.162 & 0.254 \\    \hline
    {consistency}  & 0.532 & 0.213 & 0.316  & 0.716 & 0.112 & 0.214  & 0.732 & 0.093 & 0.069 \\    \hline
    {logit}  & 0.645 & 0.048 & 0.222  & 0.762 & 0.066 & 0.173  & 0.909 & 0.220 & 0.106 \\    \hline
    \end{tabular}
    \end{tabular}
    }
    }
\caption{Calibration performance of three mainstream confidence estimation methods. The closer to orange, the better the performance; the closer to green, the worse the performance.}
\label{table: conf 3 method}
\end{table*}

% We find that existing confidence estimation methods exhibit a poor calibration performance. 
% We divided the scenarios based on the performance of the classification task and the probability distribution of the model's output, exploring the reasons for the variation in the performance of uncertainty estimation methods due to changes in scenarios.
% Different models exhibit different and even opposite trends when adjusting decoding parameters (temperature and top p sampling). No specific prompt pattern shows superior uncertainty calibration performance.

% We point out the performance of uncertainty estimation methods varies in different scenarios in Sec.~\ref{sec: different scenarios}. We analyze the reasons why each uncertainty estimation method performs best in a specific scenario in Sec.~\ref{sec: auc} - Sec.~\ref{sec: consistency-best}.
% We point out the common drawbacks of these uncertainty estimation methods in Sec.~\ref{sec: common drawbacks}. 
% Then we discuss the effects of prompt patterns in Sec.~\ref{sec: effect-prompt pattern} and the effects of temperature and top p sampling in Sec.~\ref{sec: effect-temperature}.
% \vspace{-2mm}

% \subsection{Calibration varies in different scenarios}
% \label{sec: different scenarios}

% Table~\ref{table: conf 3 method} shows the calibration performance of three mainstream confidence estimation methods. The closer to orange, the better the performance. The closer to green, the worse the performance. 

\begin{figure}
    \centering \includegraphics{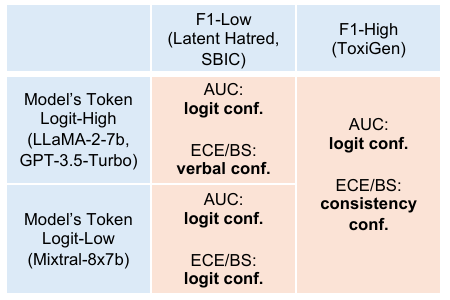}
    \vspace{-4mm}
    \caption{The best-performing uncertainty estimation method in different scenarios categorized by the model's output token logit and primary classification performance. Logit-based confidence scores achieve the best AUC in all scenarios, while the ECE for each method varies across scenarios.}
    \vspace{-4mm}
    \label{fig: conf-scenario}
\end{figure}

The calibration results in Table~\ref{table: conf 3 method} can be categorized into several scenarios as shown in Fig.~\ref{fig: conf-scenario}.
The scenarios are divided based on the primary classification performance and the logit distribution of model output tokens.

% The primary classification performance varies across datasets due to different dataset complexity. 
% LLMs exhibit low accuracy on the challenging Latent Hatred and SBIC datasets and high accuracy on the simple Toxigen dataset (as shown in Fig.~\ref{fig: oversensitive}). 
% For the logit of the output,  LLaMA-2-7b and GPT-3.5-turbo tend to assign a high logit to their output, while Mixtral-8x7b tends to assign a low logit.

The uncertainty estimation method that yields the best calibration performance is highlighted in each scenario. We find that the logit-based confidence achieves the highest AUC across all the datasets and LLMs. However, the uncertainty estimation method achieving the highest ECE/BS varies depending on the scenario. 

\begin{figure*}
	\centering
        \begin{minipage}[t]{1\linewidth}
		\centering
            \includegraphics[width=6.3in]{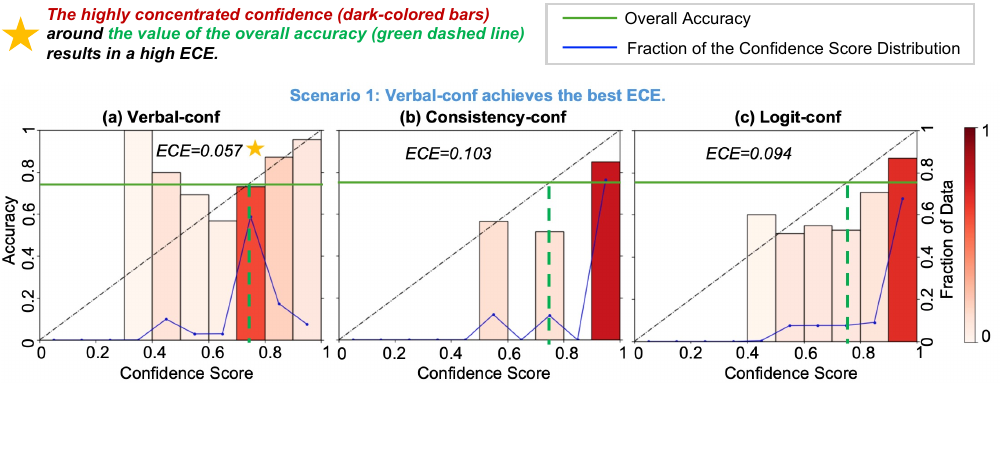}
            \vspace{-15mm}
            \caption{
            The ECE performance of LLaMA-2-7b on the SBIC dataset shows that the verbal-based confidence is mainly concentrated in the 70\%-80\% range, around the overall accuracy of 77\%, thus achieving the best ECE.}
            \label{fig: ece-llama-implicit-3method}
            \vspace{5mm}
	\end{minipage}
        \\
        \begin{minipage}[t]{1\linewidth}
		\centering
            \includegraphics[width=6.3in]
        {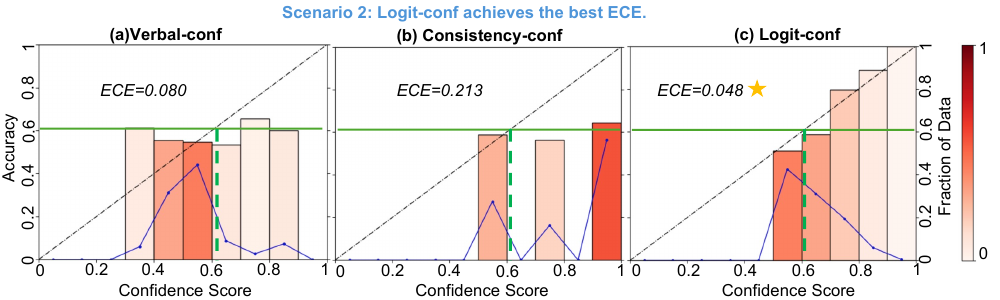}
            % \vspace{-12mm}
            \caption{
            The ECE performance of Mixtral-8x7b on the Latent Hatred dataset shows that the logit-based confidence is mainly concentrated in the 50\%-70\% range, around the overall accuracy of 61\%, thus achieving the best ECE.
            }
            \label{fig: ece-mixtral-implicit-3method}
            \vspace{5mm}
	\end{minipage}
        \\
	\begin{minipage}[t]{1\linewidth}
		\centering
            \includegraphics[width=6.3in]{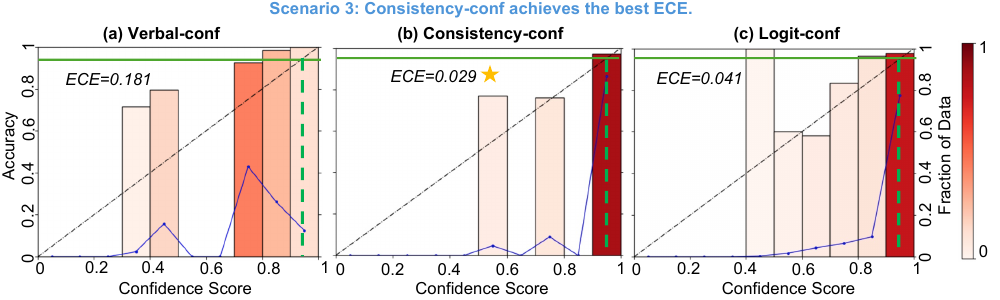}
            % \vspace{-12mm}
            \caption{
            % prob distribution and ece llama toxigen 3 method
            The ECE performance of LLaMA-2-7b on the ToxiGen shows that the consistency-based confidence is mainly concentrated in the 90\%-100\% range, around the overall accuracy of 92\%, thus achieving the best ECE.
            }
            \label{fig: ece-llama-toxigen-3method}
	\end{minipage}
\end{figure*}

We also find that the confidence distribution exhibits extremes, whether it's overly conservative or overly confident (Fig.~\ref{fig: ece-llama-implicit-3method},~\ref{fig: ece-mixtral-implicit-3method},~\ref{fig: ece-llama-toxigen-3method}.
The verbal-based confidence estimation method typically produces a relatively conservative confidence score. The consistency-based method usually demonstrates a very high confidence score. 
The distribution of logit-based confidence varies depending on the model, with GPT-3.5-Turbo and LLaMA-2-7b tending to generate high logits, while Mixtral-8x7b tends to produce conservative logits.

Our subsequent analysis reveals that the calibration performance significantly depends on the primary classification performance due to the highly concentrated confidence distribution.

\vspace{-2mm}
\begin{figure}
    \centering
\includegraphics{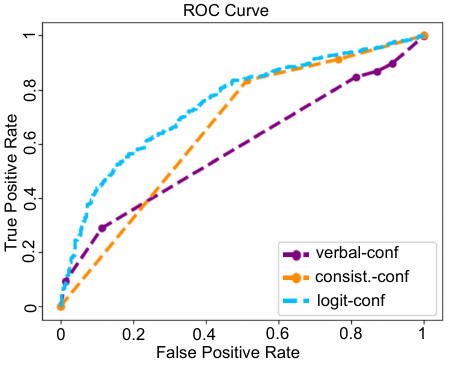}
    \vspace{-8mm}
    \caption{
    % AUC of llama implicit 3 method
    % The comparison of the ROC curve. (LLaMA-2-7b on Hatred)
    The comparison of the ROC curve.}
    \vspace{-2mm}
    \label{fig: auc-llama-implicit-3method}
\end{figure}

\begin{figure}[!ht]
    \centering
\includegraphics{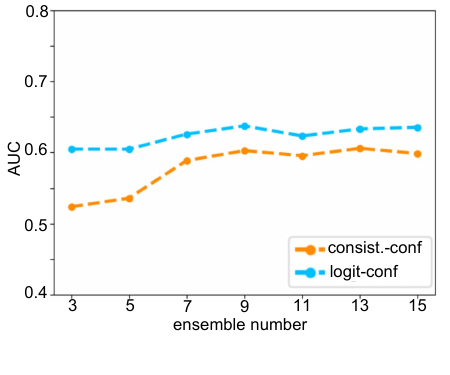}
    \vspace{-12mm}
    \caption{The figure showcases the relationship between AUC and the ensemble number.}
    \vspace{-2mm}
    \label{fig: auc-increase-ensemble-num}
\end{figure}
\vspace{-2mm}
\subsection{The logit-based confidence achieves the highest AUC in all scenarios}
\label{sec: auc}

The logit-based method performs better in AUC than both the verbal-based method and the consistency-based method in all scenarios. We compare the ROC curve composed of false positive rate (FPR) and true positive rate (TPR) in Fig.~\ref{fig: auc-llama-implicit-3method} for LLaMA-2-7b on Latent Hatred.

\textbf{The AUC of the verbal-based method is lower than the logit-based method due to the conservative verbalized confidence.} 
We observe that, at the same confidence threshold, the FPR of the verbal-based method and the logit-based method are similar. However, the confidence distribution of the logit-based method is more overconfident, leading to a larger TPR, resulting in a higher ROC curve. 
% For example, when the FPR is 0.81, the TPR for verbal-based method is 0.84 and the TPR for logit-based method is 0.94, causing the logit curve to be higher than the verbal curve.

% \textbf{The logit-based method outperforms the consistency-based method in AUC because the consistency-based confidence is too discrete.}
\textbf{The discreteness of confidence score in the consistency-based method leads to a reduced AUC.}
The consistency score in the consistency-based method is limited by the ensemble number, mainly concentrated on a few discrete values (for example, 3/5, 4/5, 5/5), leading to fewer confidence thresholds considered when calculating AUC. In contrast, the logit-based method's confidence score covers many continuous values, enabling more values of the confidence threshold for the calculation of AUC. 

As shown in Fig.~\ref{fig: auc-llama-implicit-3method}, 
% when the fpr is higer than 0.51, the points on the ROC curves for both methods are very close. 
% Both methods reach the point (fpr 0.51, tpr 0.83) with the threshold of 0.8 for logit-based method and threshold 1 for consistency-based method. 
% However, when the logit confidence is larger than 0.8, there are corresponding fpr values lower than 0.5, while the absence of consistency confidence between 0.8 and 1 has resulted in the omission of the corresponding section with fpr below 0.5.
The discrete points obtained by the consistency-based method on the ROC curve are very close to those on the curve of the logit-based method. However, the absence of a consistency-based confidence score between 0.8 and 1 results in the omission of the corresponding section with an FPR below 0.5.
This indicates that the discreteness of confidence values in the consistency-based method limits its ability to express uncertainty. 

Based on the findings, we increase the ensemble number from 3 to 13 (changing demonstrations in the prompt in each inference to conduct the ensemble), the AUC gradually increases and tends to stabilize (Fig.~\ref{fig: auc-increase-ensemble-num}). It indicates that increasing the number of ensemble sources can mitigate the gap but is still lower than the logit-based method.

% However, the advantage of the logit-based method in AUC does not always manifest in ECE. We discuss the reasons for the variations in ECE in the following sections.

These findings suggest that relying solely on AUC for comparison among different kinds of uncertainty estimation methods is insufficient. 
The variations in ECE further support this conclusion. 
Next, we delve into a detailed analysis of the reasons for the ECE variations.

\vspace{-2mm}
\subsection{Calibration heavily relies on the primary task due to concentrated confidence}
\label{sec: verbal-best}

The uncertainty estimation method achieving the highest ECE varies depending on the scenario. To demonstrate the underlying reasons, we provide an example for each scenario regarding the ECE performance in Fig.~\ref{fig: ece-llama-implicit-3method} - Fig.~\ref{fig: ece-llama-toxigen-3method}.
The x-axis represents confidence scores, binned at intervals of 0.1. The left y-axis represents accuracy. The height of each bar represents the accuracy of the corresponding bin. Darker bars indicate a higher volume of data falling within that bin. The color of the bars is quantified by a blue line, corresponding to the proportion of data on the right y-axis. The green line represents the overall accuracy of the model on the dataset.

Recall that ECE measures the gap between confidence and accuracy. As shown in Fig.~\ref{fig: ece-llama-implicit-3method},~\ref{fig: ece-mixtral-implicit-3method},~\ref{fig: ece-llama-toxigen-3method}, the highly concentrated confidence around the value of the overall accuracy results in a high ECE. 

\textbf{When does the verbal-based method achieve the highest ECE?} In cases where the performance of the primary classification task is poor and the model's token logit is high (LLaMA-2-7b on the Latent Hatred and SBIC datasets, GPT-3.5-turbo on the Latent Hatred dataset), the verbal-based method achieved nearly the best ECE and BS. 
% For example, on the implicit dataset, the ECE of the verbal-based method was the best at 0.081, significantly outperforming the logit-based method with an ECE of 0.154. 

% The confidence distributions concentrate in the interval greater than 0.9 for both the logit-based method and the consistency-based method.
% However, the accuracy within this interval is not very high, leading to confidence levels lower than the expected accuracy, 

That is because the logit-based method and the consistency-based method exhibit overconfidence while the verbal-based method provides a more conservative estimate of confidence. 
The majority of confidence scores are above 0.9 for both logit-based and consistency-based methods, yet the accuracy of the task remains low, resulting in under-calibrated errors (Fig.~\ref{fig: ece-llama-implicit-3method}). In contrast, the verbalized confidence concentrates in the range of 0.7-0.8, close to the accuracy, resulting in smaller calibration errors.

\textbf{When does the logit-based method achieve the highest ECE?}
In cases where the performance of the primary classification task is poor and the model's token logit is not generally too high (Mixtral-8x7b on the Latent Hatred and SBIC datasets), the logit-based method achieves the best calibration performance. 

That is because the conservative logit is already well-calibrated so the verbal-based method loses its advantage. 
% As shown in Fig.~\ref{fig: ece-mixtral-implicit-3method}, unlike other models that have distributions in higher score ranges, the logit of the mixtral model is mainly distributed in the 0.5-0.6 range, closely resembling the accuracy and exhibiting good calibration ability.
As shown in Fig.~\ref{fig: ece-mixtral-implicit-3method}, unlike other models that output a high token logit, the logit of the Mixtral-8x7b model is predominantly concentrated between 0.5 and 0.6, closely resembling the mediocre accuracy and exhibiting good calibration ability.
However, the consistency-based method still maintains excessive confidence, resulting in poor calibration.

% (4) consistency-based method在所有模型上表现都不好。和logit相比，因为都是overconfident，所以在ece上差距不大,但auc低于logit-based method. 这是因为consistency score受multi-inference次数的影响，主要集中的几个离散值上（3/5，4/5，5/5），导致计算auc时confidence threshold取的比较少（图4a）。而logit方法的confidence score能取到很多连续值(图4b)。confidence的分布影响了auc的计算。例如在图4c中，当threshold取0.8时，两种方法的roc曲线上的点很接近，然而当logit conf在threshold取0.99至0.8时，fpr从0.01到0.64都有，而consistency方法这部分是缺失的。这说明consistenty-based method的confidence值的离散性导致了uncertainty的表达能力受限。

\textbf{When does the consistency-based method achieve the highest ECE?}
In cases where the classification has high accuracy (all models on the ToxiGen dataset), the consistency-based method achieves the best ECE.

This is because, on simple datasets, the task accuracy is very high, so the consistency-based method with high confidence tends to be closer to the high accuracy (Fig.~\ref{fig: ece-llama-toxigen-3method}).
However, the verbal-based method maintains a conservative confidence which is lower than the high accuracy, thus leading to over-calibration. For the logit-based method, when the token logit is high (LLaMA2-7b-chat, GPT-3.5-turbo), the ECE is very similar to the consistency method. However, the logit-based method for Mixtral-8x7b has a quite lower ECE because of its conservative token logit.

% This indicates that verbal cannot effectively differentiate confidence across different datasets, as verbal confidence is influenced by demonstrations.

% 应该加一组实验，探究demonstration对verbal的影响。分别是zero-shot, 70(完成), 95

\vspace{-2mm}
\subsection{Common drawbacks}
\label{sec: common drawbacks}
% 我们发现所有的不确定性估计方法都不能良好地估计出不确定性。
These three methods are unable to effectively estimate the confidence of the answers.
% (1) 正确和错误的overlap应该小，否则说明区分不开正确错误

% 不管如何变换数据集，每种confidence score都集中在固定的区间。对llama来说，verbalized confidence都主要集中在0.7-0.8, consistency-based method和logit-based method都集中在0.9-1. 
% The confidence is highly concentrated, leading to calibration performance depending on the primary classification performance.
The calibration performance significantly depends on the primary classification performance.
No matter whether the dataset is easy or challenging, the confidence scores of each method are always concentrated in a fixed range. 
% For llama, verbalized confidence is mainly concentrated in the range of 0.7-0.8, while consistency-based methods and logit-based methods are both concentrated in the range of 0.9-1. 
Consequently, methods concentrated in low-confidence ranges perform well on challenging tasks, while those concentrated in high-confidence ranges excel in simpler tasks.
This is also why different uncertainty estimation methods achieve the best performance in different scenarios.

\begin{figure}
    \centering
\includegraphics{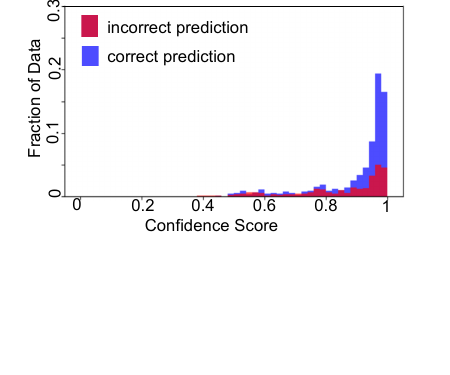}
    \vspace{-30mm}
    \caption{The confidence distribution of correctly classified and misclassified cases.}
    \vspace{-2mm}
    \label{fig: overlap-conf}
\end{figure}

% \begin{figure}
%     \centering
%     \includegraphics{_fig/experiment/distribution_implicit.pdf}
%     \vspace{-20mm}
%     \caption{The confidence distributions of correctly classified and misclassified cases overlap significantly, indicating that the existing uncertainty estimation methods can not effectively differentiate confidence in predictions.}
%     \label{fig: confidence discriminability}
% \end{figure}

Moreover, these methods struggle to distinguish the confidence between incorrectly predicted and correctly predicted instances. An ideal confidence estimation method should have high confidence for correctly predicted data and low confidence for incorrectly predicted data. However, 
% Fig.~\ref{fig: confidence discriminability} 
Fig.~\ref{fig: overlap-conf} 
shows that confidence distributions of correctly classified and misclassified cases overlap significantly, indicating the poor ability of uncertainty estimation.

\vspace{-2mm}
\subsection{Effects of prompt patterns on calibration}
\label{sec: effect-prompt pattern}
Table~\ref{table: llama prompt format}, Table~\ref{table: mixtral prompt format}, and Table~\ref{table: gpt35 prompt format} (in Appendix~\ref{sec: calibration-prompt pattern}) show that the performance of different prompts varies in calibration. No prompt consistently performs better. 
% Across different models and datasets, any given prompt can achieve either the best or the worst performance. 
% gpt35在implicit数据集上auc最好的是mask，在sbic数据集上auc最好的是qa,ece最好的是mask, toxigen上最好的是choice,ece最好的是qa和target
% llama2在implicit数据集上auc最好的是mask,ece,bs最好的是choice，sbic上auc,ece,bs最好的都是choice,toxigen上auc最好的是choice，ece最好的是qa,bs最好的是target.
The ensemble of responses obtained from different prompt patterns shows a relatively better overall performance. 
This may be because different prompt patterns inspire the model to infer results along different paths, and aggregating such results better reflects the model's confidence.

% The ensemble result derived from aggregating diverse results better reflects the model's confidence.

% This might be attributed to the diverse outcomes generated by different prompts. 
% Even with diverse results, the consistent answers indicate that the model is more certain and performs better in such diverse scenarios.

% \textbf{Effects of demonstration.}
% We also explore the potential of ensemble method changing demonstrations. We fix the prompt format as vanilla QA and randomly change balanced demonstrations in each round of single inference. With an increase in the number of rounds, there is a slight improvement in the model's results, but it is not as effective as the ensemble method based on prompts. 

\vspace{-2mm}
\subsection{Effects of the temperature and sampling}
\label{sec: effect-temperature}

% Fig.~\ref{fig: temperature} shows the calibration performance when changing temperature and top p sampling of the decoder. 

We find that due to differences in the output token logit distribution, different models exhibit varying sensitivities to temperature and sampling parameters, sometimes even in opposite trends.

\begin{figure}
	\centering
        \begin{minipage}[t]{1\linewidth}
		\centering
            \includegraphics{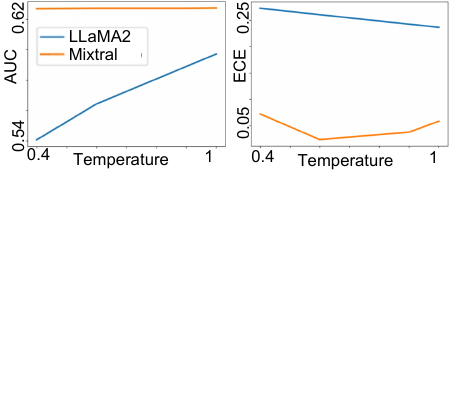}
            \vspace{-45mm}
            \caption{Calibration performance with varying temperature. LLaMA-2-7b and Mixtral-8x7b show different tends.}
            \label{fig: temperature}
            % \vspace{-800mm}
	\end{minipage}
        \\
        \vspace{-23mm}
        \begin{minipage}[t]{1\linewidth}
		\centering
            \includegraphics{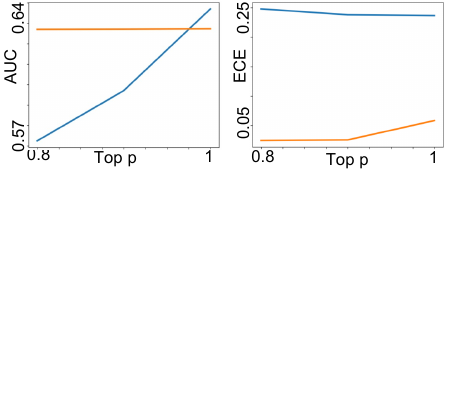}
            \vspace{-45mm}
            \caption{Calibration performance with varying top p sampling. LLaMA-2-7b and Mixtral-8x7b show different trends.}
            \label{fig: top p}
	\end{minipage}
        \vspace{-32mm}
\end{figure}

As shown in Fig.~\ref{fig: temperature}, 
as the temperature increases, The AUC of LLaMA-2-7b increases while the AUC of Mixtral-8x7b decreases. When the temperature varies between 0.6 and 1, LLaMA-2-7b and Mixtral-8x7b exhibit opposite trends in ECE.
Results on changing the top p sampling show the same findings (Fig.~\ref{fig: top p}). 
This is because the output token logit of LLaMA-2-7b is overconfident, whereas Mixtral-8x7b exhibits a more cautious level of confidence. 
As the temperature increases, the logit distribution of Mixtral-8x7b becomes sharper, leading to over-calibration. On the other hand, the logit distribution of LLaMA-2-7b becomes smoother, enhancing its ability to differentiate confidence levels. See Appendix~\ref{sec: analysis temperature} for details.

\section{Future Work}
The insights we've gained for future work include two aspects:

Firstly, future optimizations should aim to avoid misjudging benign expressions, especially those containing sensitive groups. Recent efforts aim to ensure that LLMs do not generate harmful output toward sensitive groups. However, it is unknown whether LLMs might veer towards another extreme. Imagine if all discussions about black people were filtered out on social media platforms; that would be unfair as well.

Secondly, future optimization should aim to avoid excessive concentration of confidence and should take into account varying levels of task complexity. We unveil factors influencing calibration: calibration performance significantly relies on the primary task performance due to extremely concentrated confidence.

Some future research pathways and ideas include:
(1) Integrating correction for oversensitivity during RLHF. In addition to identifying harmful content during RLHF, humans should guide LLMs to recognize challenging benign expressions involving sensitive groups but not malicious ones.
(2) Improving the reflection of confidence in logits during LLM training or inference. Optimization of the decoding process, SFT loss, or post-processing of logits can be explored to prevent the concentration of logits within a fixed range.
(3) Considering tasks of varying complexity in LLM calibration optimization. Developing uncertainty estimation methods that are adaptable to different task complexities is imperative.

\vspace{-2mm}
\section{Related Work}

\paragraph{Implicit Hate Speech Detection}
% With the proliferation of online media and user-generated content, the prevalence of hate speech has increased significantly in online platforms.
% Given the enormous volume of online posts, manually moderating all content is impractical.
% Researchers have developed various hate speech detection models, encompassing lexicon-based approaches \cite{gitari2015lexicon} and neural network models \cite{zhang2018detecting}. 
Hate speech inflicts significant harm on specific communities.
Researchers have developed various hate speech detection models \cite{gitari2015lexicon, zhang2018detecting}.
Recently, there has been a growing interest among researchers in addressing implicit hate \cite{kim2022generalizable, lin2022leveraging, ocampo2023playing}.

Certain studies have employed LLMs to generate explanations with step-by-step reasoning for detecting implicit hate speech \cite{yang2023hare}. While \citet{huang2023chatgpt} explored the quality of explanations generated by ChatGPT for detecting implicit hateful tweets, their research exclusively focused on implicit instances, neglecting non-implicit statements. In our exploration, we assess the performance of LLMs in detecting implicit hate speech, taking into account various aspects with thoughtful consideration.
We investigate whether there is an imbalance in the predictions of positive and negative classes when LLMs are employed for hate speech detection. Additionally, we thoughtfully consider various prompt patterns in our analysis.

\paragraph{Uncertainty Estimation in LLMs}
The reliable uncertainty estimation can facilitate more rational and informed decision-making. There are three mainstream methods for estimating the confidence of LLMs in their responses, particularly for cases where we can only obtain the answer and logit but not the weights. 
% The verbal-based method \cite{lin2022teaching, kadavath2022language} involves inducing LLMs to express verbalized confidence directly. Some studies utilize the consistency of multiple inferences to signify the reliability of the answer \cite{wang2022self, xiong2023can, yue2023large}. The logit-based method employs the token logit of the neural network to represent confidence \cite{Guo2017OnCO, Zhang2020MixnMatchEA, Jiang2020HowCW, Chen2022ACL}. 
They are the verbal-based method \cite{lin2022teaching, kadavath2022language, tian2023just}, the consistency-based method \cite{wang2022self, xiong2023can, yue2023large}, and the logit-based method \cite{Guo2017OnCO, Zhang2020MixnMatchEA, Jiang2020HowCW, Chen2022ACL}. 
We compare these three mainstream methods, conduct a comprehensive analysis in different scenarios, and delve into the underlying reasons. Additionally, we highlighted the shortcomings of these methods in confidence assessment for fairness.

\vspace{-2mm}
\section{Conclusion}
% The fairness and trustworthiness of LLMs have drawn widespread attention. 
% The performance of implicit hate speech detection on both the primary classification and confidence calibration is under-explored. 
The ability of LLMs in implicit hate speech detection is insufficiently examined.
In this paper, we explore the performance of LLMs in identifying implicit hate speech and the calibration of three mainstream uncertainty estimation methods. 
We find that LLMs have fallen into two extremes, excessively focusing on sensitive groups and exhibiting extreme confidence score distributions. These extremes result in over-sensitivity and poor calibration respectively.
These discoveries unveil new limitations of LLMs, underscoring the need for caution when optimizing models to ensure they do not veer towards extremes. 

\section{Limitations}
% Due to resource constraints, our current empirical evaluation has not been tested on a wider range of open-source and closed-source LLMs. 
% We only focused on English in hate speech detection in English and did not test the hate speech detection and calibration capabilities of LLMs in multiple languages.

The datasets utilized in our experiments were all in English, thus our evaluation only encompasses the English language. We did not assess whether similar issues exist for large language models in other languages such as Chinese, French, and various other languages.

% \section*{Acknowledgements}

% This document has been adapted
% by mmmm

% Entries for the entire Anthology, followed by custom entries
% \bibliography{anthology,custom}
\bibliography{refs}

\appendix
\section{Appendix}
\label{sec:appendix}

\subsection{The design of prompt patterns}
\label{sec: appendix prompt patterns}
We show the full prompt in Table ~\ref{table: show prompt format}.

% \begin{table*}[htp]
% \begin{tabular}{p{1\textwidth}}
% % \begin{tabular}{p{\linewidth}}

% \hline
% \textbf{Vanilla QA}        \\ \hline

% Given the statement, is this statement hate speech? Please provide your answer. The answer contains Yes or No. \\
% Statement: xxx \\
% Answer: \\
% \textit{Yes} \\ 

% \hline
% \textbf{Choice QA}        \\ \hline
% Given the statement, is this statement hate speech? A: Yes. B: No. Please give your choice (A or B). \\
% Statement: xxx \\
% Answer: \\
% \textit{A} \\
% \hline
% \textbf{Cloze Test}        \\ \hline
% Given the statement, is this statement hateful or neutral? Please predict the [MASK] token, you can only answer hateful or neutral. \\
% Statement: xxx \\
% Answer: It is a [MASK] statement. \\
% \textit{hateful} \\
% \hline
% \textbf{CoT}        \\ \hline
% Is this statement hate speech? Your answer should only contain Yes or No, then give your explanation step by step. \\
% Statement: xxx \\
% Answer: \\
% \textit{Yes} \\
% \textit{Explanation: xxx} \\
% \hline
% \textbf{Multi-task with Target}        \\ \hline
% Given the statement, is this statement hate speech? Your answer should only contain Yes or No, then identify the target individual / group.
%  \\
% Statement: xxx \\
% Answer: \\
% \textit{Yes} \\
% \textit{Target: xxx} \\

% \hline
% \end{tabular}
% \caption{Input and output examples for different prompt patterns. The italicized text represents the LLM's output.}
% \label{table: show prompt format}
% \end{table*}

\begin{table*}[htp]
\begin{tabular}{p{1\textwidth}}
% \begin{tabular}{p{\linewidth}}

\hline
\textbf{Vanilla QA}        \\ \hline

Given the statement, is this statement hate speech? Please provide your answer. The answer contains \\
Yes or No. \\
Statement: xxx \\
Answer: \\
\textit{Yes} \\ 

\hline
\textbf{Choice QA}        \\ \hline
Given the statement, is this statement hate speech? A: Yes. B: No. Please give your choice (A or B). \\
Statement: xxx \\
Answer: \\
\textit{A} \\
\hline
\textbf{Cloze Test}        \\ \hline
Given the statement, is this statement hateful or neutral? Please predict the [MASK] token, you can \\
only answer hateful or neutral. \\
Statement: xxx \\
Answer: It is a [MASK] statement. \\
\textit{hateful} \\
\hline
\textbf{CoT}        \\ \hline
Is this statement hate speech? Your answer should only contain Yes or No, then give your explanation \\
step by step. \\
Statement: xxx \\
Answer: \\
\textit{Yes} \\
\textit{Explanation: xxx} \\
\hline
\textbf{Multi-task with Target}        \\ \hline
Given the statement, is this statement hate speech? Your answer should only contain Yes or No, then \\
identify the target individual / group.
 \\
Statement: xxx \\
Answer: \\
\textit{Yes} \\
\textit{Target: xxx} \\

\hline
\end{tabular}
\caption{Input and output examples for different prompt patterns. The italicized text represents the LLM's output.}
\label{table: show prompt format}
\end{table*}

\subsection{Data preprocessing}
\label{sec: appendix dataset}
% Our experiments use three implicit hate speech detection datasets to assess LLMs bias on fairness: ToxiGen \cite{hartvigsen2022toxigen}, Latent Hatred \cite{elsherief2021latent}, and SBIC (v2) \cite{sap2020socialbiasframes}. 
We have two steps to preprocess the data to ensure the quality of the evaluation. Firstly, we discard the data samples with profanity words, such as "bi*ch" and "fu*k" to further ensure that the data does not contain explicit hate words. Secondly, we sample from the test set to keep the equal data number of positive and negative class. Finally, we retain 1200 test data for Latent Hatred, 1200 test data for SBIC, and 260 test data for ToxiGen.

\subsection{Examples in different datasets}
\label{sec: oversensitive examples}
The varying degrees of over-sensitivity exhibited by the model across different datasets are attributed to the varying levels of difficulty in the data. The model displays excessive sensitivity on Latent Hatred and SBIC datasets, while it does not exhibit such over-sensitivity on the ToxiGen dataset.
That is because examples of the negative class in the dataset Latent Hatred and SBIC contain sensitive groups or sensitive topics. For example, the statement "vandals target black women Maryland home" is a description of a news event that involves a fairness issue but does not contain any hateful emotions. However, examples of the negative class in dataset ToxiGen are simple ("Muslims in the 19th century migrated from Europe and the Middle East to the U.S.") and don't contain sensitive groups or topics ("I felt so much better after confiding that in someone and realizing I'm not alone"). Therefore, the model demonstrates excessive sensitivity errors on the Latent Hatred and SBIC datasets, while not exhibiting over-sensitivity on the ToxiGen dataset.

\subsection{Classification performance of different prompt patterns}
\label{sec: classification-prompt pattern}
% Please add the following required packages to your document preamble:
% \usepackage[table,xcdraw]{xcolor}
% Beamer presentation requires \usepackage{colortbl} instead of \usepackage[table,xcdraw]{xcolor}
\begin{table*}[htp]
\begin{tabular}{llllllllll}
\hline
                            & \multicolumn{3}{l}{Latent Hatred} & \multicolumn{3}{l}{SBIC}          & \multicolumn{3}{l}{ToxiGen}       \\ \hline
                            Prompt & P      & R      & F1              & P      & R      & F1              & P      & R      & F1              \\ \hline
\multicolumn{10}{l}{GPT-3.5-Turbo}                                                                                                       \\ \hline
Choice QA                      & 0.7014 & 0.758  & 0.7286 & 0.864  & 0.8381 & 0.8508 & 1      & 0.969  & 0.9843 \\
CoT                     & 0.7347 & 0.7236 & 0.7291 & 0.8895 & 0.7917 & 0.8377          & 1      & 0.9453 & 0.9719          \\
Cloze                        & 0.738  & 0.6935 & 0.715  & 0.9093 & 0.8017 & 0.8521 & 1      & 0.9225 & 0.9597          \\
Vanilla QA                          & 0.7607 & 0.6263 & 0.687           & 0.9039 & 0.7696 & 0.8314          & 1      & 0.9457 & 0.9721 \\
Target                      & 0.7452 & 0.6566 & 0.6981          & 0.9152 & 0.755  & 0.8274          & 1      & 0.9219 & 0.9593          \\ \hline
\multicolumn{10}{l}{LLaMA-2-7B}                                                                                                          \\ \hline
\multicolumn{1}{c}{Choice QA} &
  0.6143 &
  0.6913 &
  0.6505 &
0.7122 &
0.8956 &
0.7934 &
  0.9487 &
  0.881 &
0.9136 \\
\multicolumn{1}{c}{CoT} & 0.6472 & 0.6134 & 0.6299          & 0.7548 & 0.7913 & 0.7726 & 0.9344 & 0.8837 & 0.9084 \\
\multicolumn{1}{c}{Cloze} &
0.5954 &
0.9258&
0.7248&
0.6947 &
0.931 &
0.7957 &
0.8394 &
0.9583 &
  0.8949 \\
\multicolumn{1}{c}{Vanilla QA}      & 0.6373 & 0.8038 & 0.6425          & 0.8092 & 0.7646 & 0.7863 & 0.9508 & 0.8992 & 0.9243 \\
\multicolumn{1}{c}{Target} &
0.601 &
0.8395 &
0.7005 &
0.7261 &
0.8645 &
0.7893 &
  0.937 &
  0.9297 &
0.9333 \\ \hline
\multicolumn{10}{l}{Mixtral-8x7b}                                                                                                            \\ \hline
\multicolumn{1}{c}{Choice QA}  & 0.5161 & 0.995  & 0.6796          & 0.5    & 1      & 0.6667          & 1      & 0.1628 & 0.28            \\
\multicolumn{1}{c}{CoT} & 0.6155 & 0.8124 & 0.7004          & 0.7896 & 0.8633 & 0.8248          & 0.9231 & 0.9302 & 0.9266          \\
\multicolumn{1}{c}{Cloze}    & 0.503  & 0.9983 & 0.6689          & 0.5642 & 0.9817 & 0.7165          & 0.8105 & 0.9612 & 0.8794          \\
\multicolumn{1}{c}{Vanilla QA}      & 0.5771 & 0.8342 & 0.6822          & 0.584  & 0.985  & 0.7333          & 0.9141 & 0.907  & 0.9105          \\
\multicolumn{1}{c}{Target}  & 0.6058 & 0.8107 & 0.6934          & 0.8    & 0.8333 & 0.8163          & 0.8311 & 0.9535 & 0.8881          \\ \hline

\end{tabular}
\caption{The classification performance (Precision, Recall and F1) of LLMs in hate speech detection with different prompt patterns.}
\label{table: prf}
\end{table*}
The precision, recall and F1 for classification performance can be found in tables~\ref{table: prf}.

\subsection{Calibration performance of different prompt patterns}
\label{sec: calibration-prompt pattern}
Table~\ref{table: llama prompt format}, Table~\ref{table: mixtral prompt format}, and Table~\ref{table: gpt35 prompt format} show that the performance of different prompts varies in calibration.

\begin{table*}[]
\centering
\scalebox{0.8}{
\begin{tabular}
% {ccccccccccccc}
{
    S
    P{min=0.6141, max=0.6471, first color=green, second color=orange}{}
    P{min=0.1269, max=0.2453, first color=orange, second color=green}{}
    P{min=0.2416, max=0.2955, first color=orange, second color=green}{}
    P{min=0.6928, max=0.7505, first color=green, second color=orange}{}
    P{min=0.0882, max=0.1523, first color=orange, second color=green}{}
    P{min=0.1652, max=0.2052, first color=orange, second color=green}{}
    P{min=0.8227, max=0.92, first color=green, second color=orange}{}
    P{min=0.0084, max=0.0688, first color=orange, second color=green}{}
    P{min=0.0471, max=0.0776, first color=orange, second color=green}{}
  }
    \hline
\multicolumn{1}{c}{}   & \multicolumn{3}{c}{Latent Hatred}      & \multicolumn{3}{c}{SBIC}          & \multicolumn{3}{c}{Toxigen}      \\ \hline
\multicolumn{1}{c}{Prompt}  & \multicolumn{1}{c}{AUC}    & \multicolumn{1}{c}{ECE} & \multicolumn{1}{c}{BS}    & \multicolumn{1}{c}{AUC}  & \multicolumn{1}{c}{ECE}  & \multicolumn{1}{c}{BS} & \multicolumn{1}{c}{AUC}    & \multicolumn{1}{c}{ECE} & \multicolumn{1}{c}{BS}     \\ \hline
{Choice QA}   & 0.628 & 0.127 & 0.242 &  0.751 & 0.088 & 0.167 & 0.924 & 0.069 & 0.052 \\ \hline
{CoT}  & 0.631 & 0.215 & 0.279 & 0.736 & 0.139 & 0.191 & 0.902 & 0.037 & 0.056 \\ \hline
{Cloze}     & 0.647 & 0.245 & 0.296 &  0.696 & 0.152 & 0.205 & 0.877 & 0.013 & 0.078 \\ \hline
{Vanilla QA}       & 0.637 & 0.236 & 0.287 & 0.693 & 0.129 & 0.186 & 0.823 & 0.008 & 0.059 \\ \hline
{Target}  & 0.614 & 0.235 & 0.294 & 0.720 & 0.141 & 0.196 & 0.880 & 0.026 & 0.048 \\ \hline
{Ensemble} & 0.637 & 0.154 & 0.244 & 0.749 & 0.094 & 0.165 & 0.889 & 0.041 & 0.047 \\ \hline
\end{tabular}
}
\caption{The calibration performance of LLaMA-2-7B with different prompt patterns.}
\label{table: llama prompt format}
\end{table*}

\begin{table*}[]
\centering
\scalebox{0.8}{
\begin{tabular}
% {ccccccccccccc}
{
    S
    P{min=0.6271, max=0.7542, first color=green, second color=orange}{}
    P{min=0.0257, max=0.2419, first color=orange, second color=green}{}
    P{min=0.2093, max=0.2768, first color=orange, second color=green}{}
    P{min=0.7282, max=0.8363, first color=green, second color=orange}{}
    P{min=0.0663, max=0.1839, first color=orange, second color=green}{}
    P{min=0.1364, max=0.2459, first color=orange, second color=green}{}
    P{min=0.8418, max=0.9369, first color=green, second color=orange}{}
    P{min=0.1016, max=0.2726, first color=orange, second color=green}{}
    P{min=0.0652, max=0.2106, first color=orange, second color=green}{}
}
  \hline
\multicolumn{1}{c}{}   & \multicolumn{3}{c}{Latent Hatred}      & \multicolumn{3}{c}{SBIC}          & \multicolumn{3}{c}{Toxigen}       \\ \hline
\multicolumn{1}{c}{Prompt}  & \multicolumn{1}{c}{AUC}    & \multicolumn{1}{c}{ECE} & \multicolumn{1}{c}{BS}    & \multicolumn{1}{c}{AUC}  & \multicolumn{1}{c}{ECE}  & \multicolumn{1}{c}{BS} & \multicolumn{1}{c}{AUC}    & \multicolumn{1}{c}{ECE} & \multicolumn{1}{c}{BS}     \\ \hline
{Choice QA}   & 0.710 & 0.095 & 0.237 & 0.836 & 0.184 & 0.246 & 0.842 & 0.273 & 0.211 \\ \hline
{CoT}  & 0.700 & 0.039 & 0.209 & 0.787 & 0.091 & 0.136 & 0.927 & 0.127 & 0.065 \\ \hline
{Cloze}     & 0.668 & 0.242 & 0.277 & 0.770 & 0.157 & 0.217 & 0.879 & 0.102 & 0.098 \\ \hline
{Vanilla QA}       & 0.682 & 0.026 & 0.228 & 0.728 & 0.094 & 0.204 & 0.911 & 0.265 & 0.138 \\ \hline
{Target}   & 0.693 & 0.028 & 0.210 & 0.781 & 0.100 & 0.142 & 0.888 & 0.154 & 0.104 \\ \hline
{Ensemble} & 0.710 & 0.048 & 0.222 & 0.762 & 0.066 & 0.173 & 0.924 & 0.220 & 0.106 \\ \hline
\end{tabular}
}
\caption{The calibration performance of Mixtral-8x7b with different prompt patterns.}
\label{table: mixtral prompt format}
\end{table*}
\begin{table*}[]
\centering
\scalebox{0.8}{
\begin{tabular}
% {ccccccccccccc}
{
    S
    P{min=0.6375, max=0.6832, first color=green, second color=orange}{}
    P{min=0.1505, max=0.1905, first color=orange, second color=green}{}
    P{min=0.2189, max=0.2483, first color=orange, second color=green}{}
    P{min=0.8004, max=0.858, first color=green, second color=orange}{}
    P{min=0.0668, max=0.0901, first color=orange, second color=green}{}
    P{min=0.1182, max=0.1405, first color=orange, second color=green}{}
    P{min=0.9106, max=0.9732, first color=green, second color=orange}{}
    P{min=0.0289, max=0.0475, first color=orange, second color=green}{}
    P{min=0.0153, max=0.0303, first color=orange, second color=green}{}
  }
  \hline
\multicolumn{1}{c}{}   & \multicolumn{3}{c}{Latent Hatred}      & \multicolumn{3}{c}{SBIC}          & \multicolumn{3}{c}{Toxigen}       \\ \hline
\multicolumn{1}{c}{Prompt}  & \multicolumn{1}{c}{AUC}    & \multicolumn{1}{c}{ECE} & \multicolumn{1}{c}{BS}    & \multicolumn{1}{c}{AUC}  & \multicolumn{1}{c}{ECE}  & \multicolumn{1}{c}{BS} & \multicolumn{1}{c}{AUC}    & \multicolumn{1}{c}{ECE} & \multicolumn{1}{c}{BS}     \\ \hline
{Choice QA}    & 0.646 & 0.187 & 0.248 & 0.811 & 0.076 & 0.125 & 0.973 & 0.047 & 0.015 \\
\hline
{CoT}   & 0.654 & 0.171 & 0.229 & 0.818 & 0.083 & 0.132 & 0.911 & 0.031 & 0.026 \\
\hline
{Cloze}      & 0.683 & 0.177 & 0.232 & 0.823 & 0.069 & 0.123 & 0.963 & 0.048 & 0.024 \\
\hline
{Vanilla QA}        & 0.638 & 0.191 & 0.246 & 0.834 & 0.090 & 0.138 & 0.924 & 0.029 & 0.026 \\
\hline
{Target}    & 0.658 & 0.183 & 0.241 & 0.800 & 0.090 & 0.141 & 0.936 & 0.030 & 0.030 \\
\hline
{Ensemble}  & 0.668 & 0.151 & 0.219 & 0.858 & 0.067 & 0.118 & 0.959 & 0.045 & 0.021 \\
\end{tabular}
}
\caption{The calibration performance of GPT-3.5-Turbo with different prompt patterns.}
\label{table: gpt35 prompt format}
\end{table*}

\subsection{Analysis of the effects on the temperature}
\label{sec: analysis temperature}
\begin{figure}
    \centering
    \includegraphics{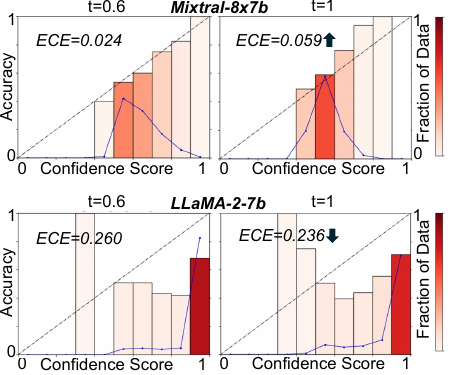}
    % \vspace{-20mm}
    \caption{The ECE performance with temperature=0.6 and temperature=1 for Mixtral-8x7b and LLaMA-2-7b. The bar’s color and blue line both represent the fraction of the data.}
    \label{fig: temperature-bar}
\end{figure}

The difference in the effect of temperature on LLaMA-2-7b and Mixtral-8x7b arises from the different logit distribution. Fig.~\ref{fig: temperature-bar} shows the ECE performance for the logit-based uncertainty estimation method with different temperatures on the Latent Hatred dataset. The confidence score for the logit-based method is the logit for the output token. 
The logit distribution of Mixtral-8x7b is primarily concentrated between 0.5 and 0.7, while LLaMA's logit is mainly distributed between 0.9 and 1.0. This indicates that LLaMA-2-7b is overconfident, whereas Mixtral-8x7b exhibits a more cautious level of confidence. 
As the temperature increases, the logits for both models become more conservative. 
Thus, the logit distribution of Mixtral-8x7b becomes sharper, leading to over-calibration. 
On the other hand, the logit distribution of LLaMA-2-7b becomes smoother, enhancing its ability to differentiate confidence levels.

\end{document}